\documentclass{article}

\usepackage[preprint]{corl_2026} 

\usepackage{amsmath}
\usepackage{amsfonts}  
\usepackage{enumitem}

\usepackage{algorithm}
\usepackage{algorithmic}

\usepackage{graphicx}
\usepackage{wrapfig}

\usepackage{booktabs}
\usepackage{multirow}
\usepackage[table]{xcolor}

\newcommand{\citec}[1]{~\cite{#1}}
\title{
ReGIL: Retrieval-Guided Imitation Learning from a Single Demonstration
}
\author{
  Yuying Zhang, Francesco Verdoja, Wenyan Yang, Ville Kyrki\\
  School of Electrical Engineering\\
  Aalto University,
  Finland\\
  \texttt{\{firstname.surname\}@aalto.fi} \\
}

\begin{document}
\maketitle

\begin{abstract}
Learning robot manipulation policies with deep neural networks from a single demonstration remains highly challenging, as even small deviations from the demonstrated trajectory can quickly compound into failure, while collecting substantial online interaction data is costly. We propose ReGIL, a retrieval-guided imitation learning framework that treats a single demonstration as an external memory. ReGIL repeatedly queries this static memory throughout training to simultaneously guide exploration, generate the regularization buffer, and construct rewards. Specifically, it computes rewards through local temporal alignment between the current trajectory and the retrieved segment, providing step-wise and informative feedback for policy improvement. We evaluate ReGIL on robotic manipulation tasks from the LIBERO and Meta-World benchmarks under the single demonstration setting. ReGIL outperforms prior baselines in both success rate and training efficiency. In real-robot experiments, using only one demonstration and less than one hour of online training, ReGIL achieves over 75\% success rate across three manipulation tasks with randomness in both initial robot pose and target position. These results demonstrate that leveraging the single demonstration as reusable memory can provide more than static supervision for efficient robot learning. More details can be found on our website: https://regil2026.github.io/

\end{abstract}


\section{Introduction}
\label{sec:introduction}

Learning complex manipulation skills from a single demonstration remains a fundamental challenge in robotics. However, pure offline learning is inherently challenging due to severe distribution shift, leading to compounding errors when the agent deviates from expert trajectories~\cite{bc_2018}. Recent retrieval-based approaches mitigate this issue by matching the current observation to the demonstrations and replaying the corresponding actions~\cite{dinobot, il_replay_search, retri_align}. While effective in some settings, all of them either replay retrieved behaviors or use them as supervision without adaptation, limiting their robustness and generalization in real-world manipulation tasks.

On the other hand, online learning allows a robot to adapt through interaction and recover from off-demonstration states, for example, through Reinforcement Learning (RL)~\cite{RL_Sutton}. However, standard RL methods are typically sample inefficient, especially for vision-based manipulation tasks~\cite{robot_survey}. With only sparse rewards and limited interaction time, uninformed exploration rarely discovers meaningful behaviors, making policy learning slow.

\begin{figure*}
    \centering
    \includegraphics[width=\textwidth]{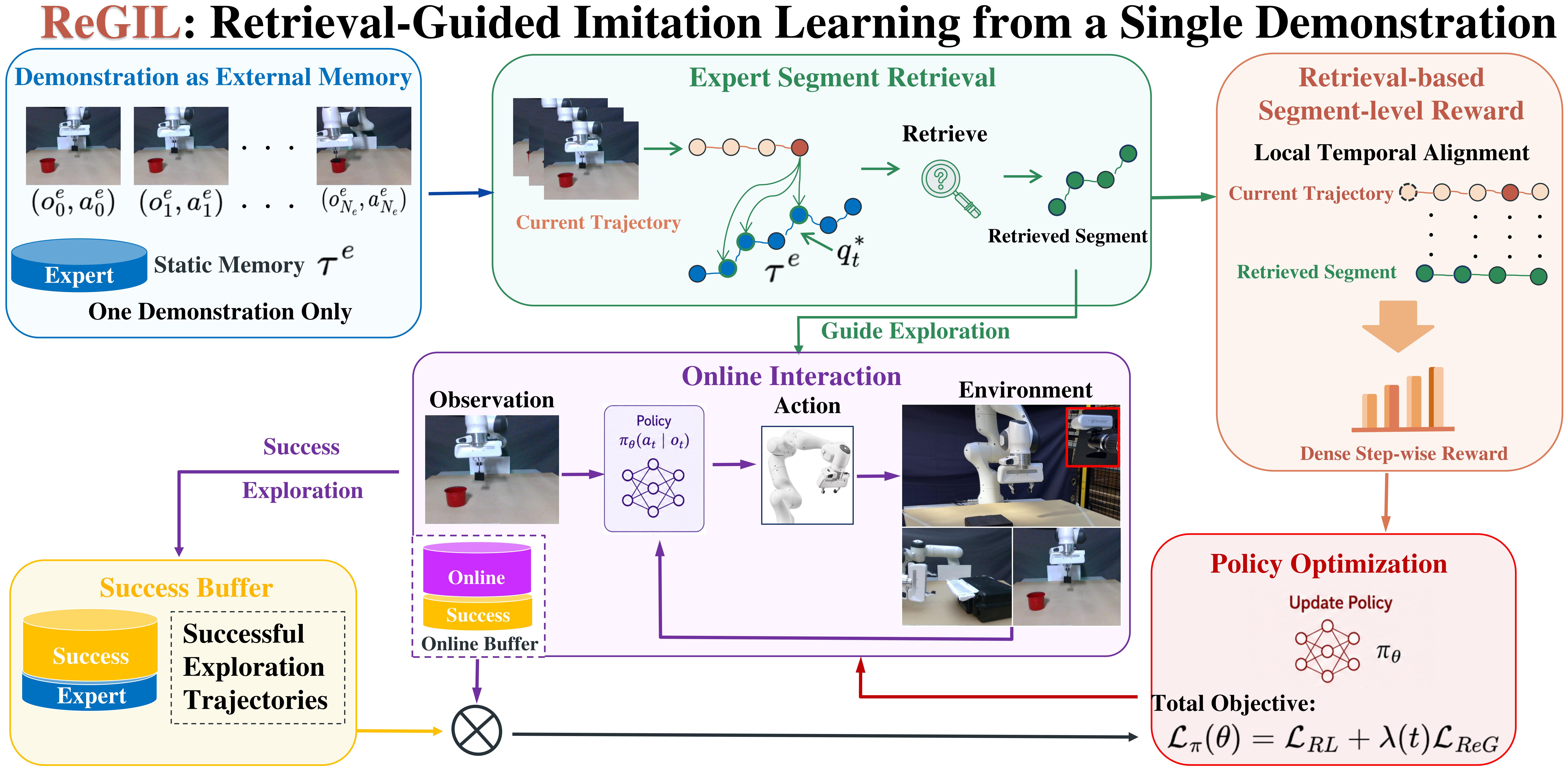}
    \vspace{-5pt}
\caption{Overview of ReGIL. ReGIL treats a single demonstration as external memory for online robot learning. At each online interaction step, the agent retrieves a task-relevant expert segment conditioned on its current trajectory. The retrieved segment is used to guide exploration and construct dense segment-level rewards for policy optimization via local temporal alignment. Successful episodes during the retrieve-guided exploration phase are added to the success buffer to increase the diversity of regularization data.
}
\label{fig:framework}
\end{figure*}

In this work, we propose ReGIL (Retrieval-Guided Imitation Learning), a framework that uses a single demonstration as persistent external memory throughout learning to inherit the strengths of both offline non-parametric retrieval and online parametric learning. The framework is shown in Fig.~\ref{fig:framework}. ReGIL continuously queries the single demonstration during interaction and leverages retrieval to support early exploration, collection of regularization data, and reward construction.

Specifically, ReGIL retrieves relevant expert segments to guide early exploration and generate meaningful trajectories. The successful rollouts obtained during this phase are stored as a success buffer, which enriches the diversity of solutions and mitigates the distribution shift introduced by online data. Leveraging regularization from the success buffer, the policy is further optimized using a retrieval-based reward proxy derived from local temporal alignment between the agent trajectory and the retrieved expert segment. This segment-level reward preserves temporal structure while providing dense and step-wise feedback during interaction.

Our main contributions are:
\begin{itemize}[nosep]

\item A retrieval-guided one-shot imitation learning framework:
ReGIL repeatedly queries the single demonstration during training to guide exploration and construct rewards, and incrementally builds a success buffer during guided exploration that increases the diversity
of regularization data beyond the original demonstration and stabilizes policy optimization;

\item A retrieval-based segment reward proxy:
ReGIL retrieves expert segments based on the current trajectory and computes rewards through local temporal alignment rather than full-trajectory matching, providing step-wise and data efficient feedback;
\item Empirical validation in simulation and real-world robotics:
ReGIL outperforms prior baselines on Meta-World and LIBERO, and learns real-world Franka Panda skills from a single demonstration with less than one hour of interaction.

\end{itemize}


\section{Related Works}
\label{sec:related_work}

\paragraph{Retrieval-based Imitation Learning} Retrieval-based imitation learning has shown strong performance in few-shot robotic manipulation by reusing expert demonstrations. Prior work uses retrieval for data augmentation \citec{retrieval_strap, retrieval_behavior, handme, ram, retri_memory}, behavior replay \citec{il_replay_search, vinn, dinobot}, subgoal generation \citec{retri_align}, or diffusion and value-based refinement \citec{retri_diffusion, retri_graph}. However, most methods treat retrieval as static supervision or directly behavior replay (e.g., VINN\citec{vinn} and DinoBot\citec{dinobot}), limiting online adaptation under disturbances. In contrast, ReGIL uses retrieval throughout training, providing exploration priors in the early stage and calculating a reward proxy during online learning. 

\paragraph{Expert Regularized Reinforcement Learning } A common strategy for improving RL sample efficiency is to incorporate expert demonstrations through online regularization or offline pretraining \citec{td3_bc, offline_RL_bc,bc_RL_jump, q_filter, bc_RL_jump_robot, ROT}. While effective, these approaches typically constrain the policy around a fixed offline demonstration dataset \citec{rl_data} or pretrained policies, making them sensitive to distribution shift when the agent encounters unseen states during interaction. RFCL\citec{RFCL} accelerates the training of the initial policy by building a reverse curriculum. In contrast, ReGIL adopts a training-free retrieval-guided exploration strategy and incrementally expands a success buffer collected online during exploration. This adaptive expanded memory increases the diversity of supervision beyond the original demonstration and mitigates the distribution shift between offline demonstrations and online interaction trajectories.

\paragraph{Trajectory Alignment and Reward Proxy }Trajectory alignment has been widely used to construct reward proxies for imitation learning when task rewards are sparse or unavailable\citec{ROT, TOT, PWIL, SIL, OTR}. Many methods formulate alignment as an optimal transport (OT) problem\citec{PWIL, SIL}, computing Wasserstein distances across entire trajectories. While effective, these global matching approaches are computationally expensive and produce reward signals that depend on the access to future trajectories. Temporal variants such as Temporally Optimal Transport (TOT) \cite{TOT} introduce masking to improve temporal consistency but still rely on global alignment. Dynamic Time Warping (DTW) and its variants explicitly model temporal correspondence and have also been used for trajectory comparison and imitation learning \cite{DTW_review, GDTW_IL, GWIL}. Recent work, such as STRAP \cite{retrieval_strap}, combines DTW with a visual foundation model (VFM) for offline retrieval-based imitation without online adaptation.

Unlike OT-based rewards, which define divergences between trajectory distributions, ReGIL computes rewards from locally retrieved trajectory segments rather than globally matching complete trajectories (Tab.~\ref{tab:comparison}). This step-wise reward depends solely on past and current observations, preserving the benefits of temporal alignment while enabling efficient dense reward computation for online learning. 


\begin{table}
\centering
\small
\caption{
Comparison of trajectory alignment reward formulations.
OT-based methods rely on global trajectory matching and require access to future trajectory information. In contrast, ReGIL performs local temporal alignment over retrieved segments, computing reward with only the trajectory history.
}
\vspace{-5pt}
\begin{tabular}{l|c cc c  }
\toprule
\textbf{Methods} & \textbf{Matching Scope} 
& \textbf{Information}
& \textbf{Reward Granularity}  
& \textbf{Matching Metric}
\\
\midrule
OT-base\cite{ROT,PWIL} 
& Full global 
& Full trajectory
& Trajectory-level
& Wasserstein

\\
TOT\citec{TOT} 
& Masked global 
& Full trajectory
& Trajectory-level
& Wasserstein

\\

\rowcolor[HTML]{FFE4E1} 
\textbf{ReGIL (Ours)} 
& \textbf{Retrieved local} 
& \textbf{Historical trajectory}
& \textbf{Segment-level}
& \textbf{S-DTW}

\\
\bottomrule
\end{tabular}
\vspace{-5pt}
\label{tab:comparison}
\end{table}

\section{Retrieval Guided Imitation Learning}
\label{sec:methods}
ReGIL addresses the limited supervision problem in one-shot robot learning by using the single expert demonstration as a queryable external memory to provide guidance throughout training. We consider a continuous-control Markov Decision Process defined by observation space $\mathcal{O}$, action space $\mathcal{A}$, and discount factor $\gamma \in (0,1)$. The agent receives only RGB image observations $o_t \in \mathcal{O}$. The single expert demonstration is denoted as $\tau^e = \{(o_0^e,a_0^e), (o_1^e,a_1^e), \dots, (o_{N_e}^e,a_{N_e}^e)\} $. A frozen visual foundation model (VFM) encoder $\Phi(\cdot)$ maps observations to an embedding space, and we denote the Euclidean distance in the embedding space by $d(o_i,o_j)=\|\Phi(o_i)-\Phi(o_j)\|_2^2$.

At each timestep, ReGIL retrieves a task-relevant expert segment based on the current observation and recent history. This retrieved segment is used to (i) guide early exploration through action replay, (ii) populate a success buffer for behavior regularization, and (iii) compute dense rewards via local temporal alignment. The retrieval process is performed once per timestep, and the retrieved information is subsequently used for both early exploration and reward calculation. ReGIL consists of three components: expert segment retrieval, retrieval-guided exploration, and retrieval-guided policy optimization. Full algorithmic details and pseudo-code are provided in Appendix~\ref{app:alg_overall}.

\subsection{Expert Segment Retrieval}
\label{sec:retrieval}
Given the current observation $o_t$, ReGIL performs a coarse-to-fine retrieval process to identify the most relevant expert segment. Instead of directly matching against the entire demonstration trajectory, we first retrieve a set of visually similar states and then select the most temporally advanced state among them to balance retrieval efficiency and avoid local optima. The final retrieved expert segment is obtained through local temporal alignment based on Subsequence Dynamic Time Warping (S-DTW)\citec{s-dtw} distance.

\paragraph{Candidate selection }Let $\mathcal{N}_k(o_t) \subseteq \{0,\dots,N_e\}$ denote the set of indices of the $k$ smallest values of $d(o_t, o^e_i)$, i.e., the top-$k$ visual nearest neighbors in the demonstration. We define the candidate segment endpoint as $q^*_t = \max\, \mathcal{N}_k(o_t)$.  This criterion ensures that the retrieved states are both visually similar and temporally progressive along the task, preventing the agent from matching locally similar but unproductive states.

\paragraph{Local temporal alignment}To reduce ambiguity from single-frame matching, we incorporate temporal consistency by aligning the recent observation history $\tau_{t-H:t} = \{o_{t-H},\dots,o_t\}$ with the expert trajectory segment $\tau^e_{0:q_t^*}=\{o_0^e,o_1^e,\dots,o_{q_t^*}^e\}$ using S-DTW, where $H$ is the history horizon. The local matching cost is defined as:
$
c(i,j)=d(o_{t-H+i-1},o_{j-1}^e),
\quad i\in\{1,\dots,H+1\},\ j\in\{1,\dots,q_t^*+1\}.
$ 
Unlike standard DTW, S-DTW\citec{s-dtw} allows the agent trajectory to align with an arbitrary segment of the expert trajectory. We compute the dynamic programming table $D \in \mathbb{R}^{(H+2)\times(q_t^*+2)}$:
\begin{equation}
D(i, j) = 
\begin{cases} 
0, & i = 0\\
\infty, & j = 0,i > 0\\
c(i, j) + \min \{ D(i-1, j-1), D(i-1, j), D(i, j-1) \}, & \text{otherwise}
\end{cases}
\label{eq:s-dtw}
\end{equation}

The optimal alignment endpoint $j_t^*$ is obtained by minimizing $D(H+1,j)$, yielding both the matched segment and the alignment cost.
\begin{equation}
   j_t^* = \arg\min_{1 \le j \le q_t^*+1} D(H+1,j)\quad 
   d_{\text{s-dtw}}(\tau_{t-H:t},\tau^e_{0:q_t^*} )= D(H+1, j_t^*)
\end{equation}

\subsection{Retrieval-Guided Exploration}
\label{sec:retrieval_exploration}
Random exploration at the early stage is highly inefficient, leading to slow convergence. Inspired by the success of the retrieve and replay policy\citec{vinn}, we therefore query the demonstration and retrieve the relevant segment to guide early exploration by executing aligned expert actions for fixed steps.

During this phase, the agent executes the aligned expert action $a_t = a^e_{j_t^*-1}$ to guide exploration toward success. Successful episodes during early exploration are added to the success buffer $\mathcal{B}_{\text{suc}}$, which later serves as a regularization dataset for policy learning to improve training stability, reduce distribution shift between the demonstration and online observation, and provide adaptive supervision beyond the single expert demonstration.

\subsection{Retrieval-Guided Policy Optimization}
For policy optimization, ReGIL interacts with the environment to maximize the retrieval-based reward with a decaying regularization.

\paragraph{Retrieval-based segment-level Reward Proxy} 
At each timestep, we compute a reward based on the alignment between the agent’s recent trajectory and the retrieved expert segment:
\begin{equation}
r_{t+1} = -\frac{1}{H+1} d_{\text{s-dtw}}( \tau_{t-H+1:t+1},\tau_{0:q_{t+1}^*}^e).
\label{eq:reward}
\end{equation}
The reward is normalized by the agent window length, providing dense, step-wise feedback that preserving temporal consistency without access to future information. To stabilize reward estimation, we use a frozen visual encoder DINOv3\citec{dinov3}, thereby decoupling reward computation from representation learning and avoiding instability caused by shifting feature distributions.

\paragraph{Optimize with Decaying Regularization} 
We adopt a TD3-style actor-critic framework~\cite{TD3} for continuous control. The critic is trained with standard Bellman updates, maintaining two Q networks $Q_{\phi_1}$ and $Q_{\phi_2}$ with definition $Q_{\min}(o,a)=\min\{Q_{\phi_1}(o,a),Q_{\phi_2}(o,a)\}$ and online replay buffer $\mathcal{B}$. The actor maximizes expected Q-values with an additional behavior cloning regularizer over trajectories in $\mathcal{B}_{\text{suc}}$ :
\begin{equation}
\mathcal{L}_{\pi}(\theta) =  - \mathbb{E}_{(o_t)\sim \mathcal{B}} \left[ Q_{\min}(o_t,\pi_\theta(o_t)) \right] 
+ \lambda(t) \cdot \mathbb{E}_{(o_i, a_i) \sim \mathcal{B}_\text{suc}} \left[-\log  \pi_\theta(a_i \mid o_i) \right]
\label{eq:policy_opt}
\end{equation}

The $\mathcal{B}_{\text{suc}}$ is initialized with $ \tau^e$ and the regularization weight $\lambda(t)$ decays over time, allowing the policy to gradually shift from imitation to reinforcement learning and ultimately improve beyond the demonstration.

\begin{figure*}[tb]
    \centering
    \includegraphics[width=\textwidth]{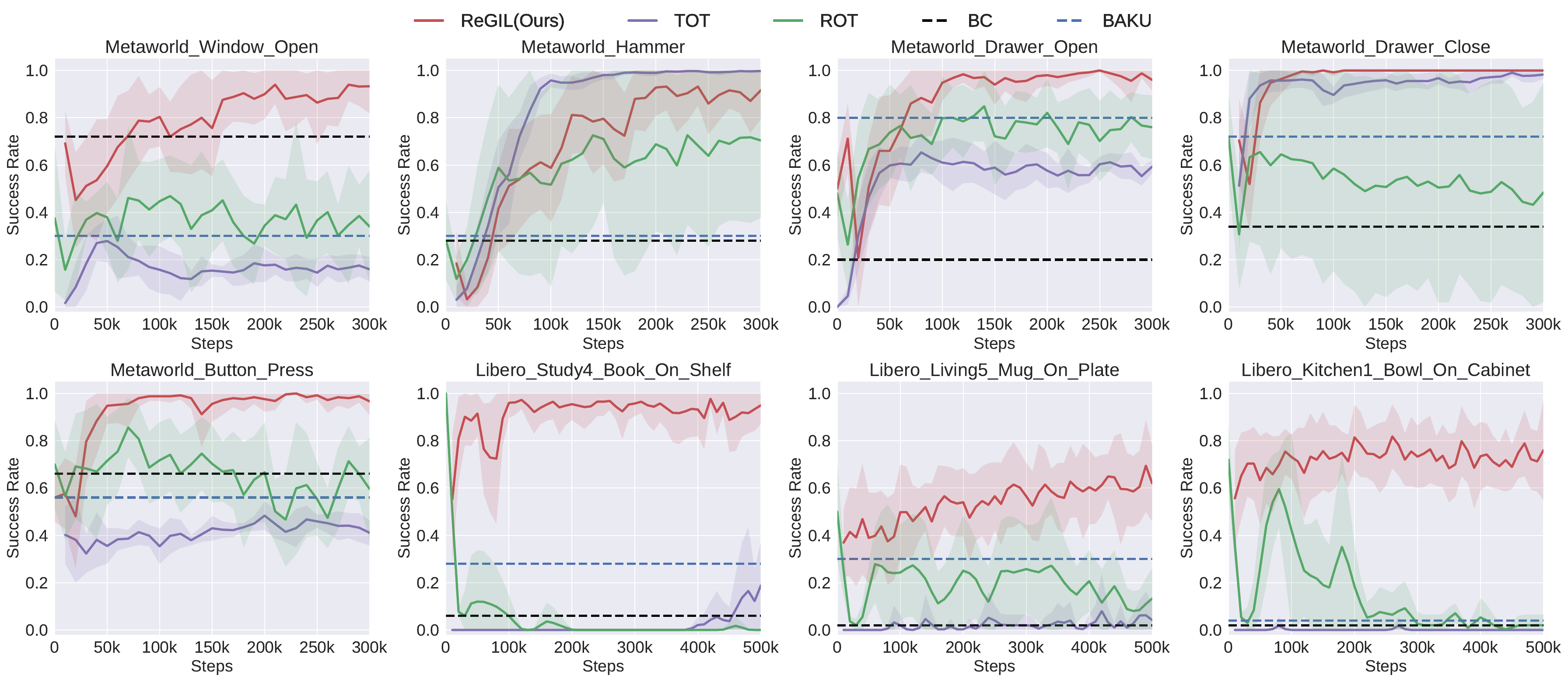} 
    \vspace{-5pt}
\caption{
Comparison to Baselines in Simulation. All methods are trained with a single demonstration on Meta-World and LIBERO. Curves show mean success rate, with shaded regions indicating $\pm$1 standard deviation across five seeds, clipped to [0, 1].
}
\label{fig:baseline}
\vspace{-5pt}
\end{figure*}

\section{Experiments}
\label{sec:result}
We evaluate ReGIL to answer the following questions:
\begin{itemize}[nosep]
\item How does ReGIL compare to state-of-the-art methods?
\item What is the contribution of each main component? 
\item How effective is the retrieval-based reward compared with alternative similarity rewards?
\item How does our method perform in real-world settings?
\end{itemize}
Further details, including visual fundamental model selection, hyperparameter sensitivity study, computational efficiency, and success buffer analysis, are provided in the Appendix~\ref{app:appendix}. 

\subsection{Experimental Setup}

\paragraph{Simulated Tasks}
We evaluate on Meta-World~\citec{metaworld} and LIBERO~\citec{libero}, covering various manipulation tasks. Policies are trained from a single demonstration per task using only RGB image observations. For Meta-World, we use the scripted expert demonstrations provided in the official benchmark implementation. For LIBERO, we use the publicly released demonstrations provided with the benchmark. Online methods are evaluated along with training over 10 rollouts, reporting the mean and standard deviation across 5 seeds. Offline methods are evaluated over 50 trials.

\paragraph{Real-world Tasks}
We deploy ReGIL on a Franka Panda robot across three tasks: (1) Reach (a fundamental positioning task), (2) Insert (requiring high precision), and (3) Open (involving rich-contact dynamics), illustrated in Fig.~\ref{fig:real_task}. Observations are $84 \times 84$ RGB images captured by a Realsense camera and actions are 4D end-effector commands $ \mathcal{A}=\mathbb{R}^4 $. Each task has a single teleoperated demonstration, and online training is limited to less than one hour, including the manual reset movement. The experiments are designed with randomized initial states, while the single demonstration only shows how to solve the task from one specific initial condition. We evaluate methods under both the \emph{fixed-target} setting (the target is placed at the same position as in the demonstration) and the \emph{random-target} setting (the target position is varied within a predefined range). Additional details of the task setup and success metrics are provided in Appendix~\ref{app:real_world}.

\subsection{Results on the Simulated Tasks}

\subsubsection{Comparison to Baselines}

We compare ReGIL against representative baselines, including the classical behavior cloning (BC), transformer-based imitation learning (BAKU)\citec{baku}, trajectory-matching inverse reinforcement learning methods (IRL) with regularization (ROT)~\cite{ROT}, and Temporal Trajectory-matching IRL (TOT)~\cite{TOT}. Additional implementation details, the baseline selection criteria, and more results are provided in Appendix~\ref{app:simulation}.
\begin{figure*}[tb]
    \centering
    \includegraphics[width=\textwidth]{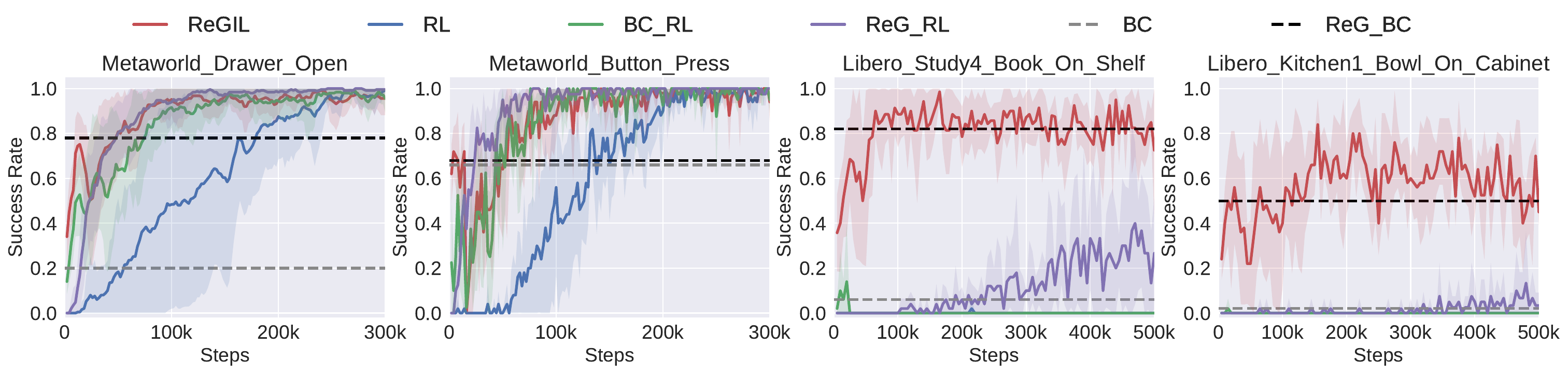}
    \vspace{-5pt}
\caption{Ablation of Components.
 We compare ReGIL with four ablated variants, including ReGIL w/o Retrieval (BC\_RL), ReGIL w/o Regularization (ReG\_RL), RL with Random Exploration (RL), and ReGIL w/o  Optimization (ReG\_BC). Results show the mean success rate across 5 random seeds,  with the shaded region representing $\pm$1 standard deviation, clipped to the range [0,1]. 
}
\vspace{-5pt}
\label{fig:ablation_com}
\end{figure*}

Fig.~\ref{fig:baseline} compares ReGIL with representative baselines on Meta-World and LIBERO. Overall, ReGIL learns faster and achieves higher final success rates on most tasks, except for the hammer task where the smaller success buffer (approximately one-third of the samples compared to other tasks) limits the diversity of successful experiences available for regularization. Compared with supervised learning methods such as BC and BAKU, ReGIL shows clear improvements, particularly on long-horizon LIBERO tasks. This highlights the difficulty of learning robust policies from a single demonstration by offline imitation learning. Compared with trajectory-matching methods such as ROT and TOT, ReGIL reaches higher success rates with fewer environment interactions, suggesting that the retrieval-based framework provides a stronger learning signal. 

\subsubsection{Ablation Studies}

\paragraph{Ablation study on each component}  ReGIL framework consists of three main components: (1) the retrieval exploration, (2) behavior regularization (BC) that keeps the learned policy close to the success buffer $\mathcal{B}_{\text{suc}}$, and (3) reinforcement learning (RL) further fine-tunes the policy. To better understand the contribution of each component in ReGIL, we conduct ablation studies based on the following ablated variants: ReGIL w/o Retrieval (BC\_RL), ReGIL w/o Regularization (ReG\_RL), ReGIL w/o RL (ReG\_BC), and ReGIL w/o both Retrieval and Regularization (RL). The ablation experiments are conducted on four tasks, two from the Metaworld benchmark and two from the LIBERO task suite. The results are shown in Fig.~\ref{fig:ablation_com}. Overall, both retrieval and BC regularization significantly improve learning efficiency, with retrieval playing a more critical role in long-horizon LIBERO tasks. Without retrieval (BC\_RL), the policy fails on LIBERO and learns more slowly on MetaWorld, indicating that RL with a single demonstration provides an insufficient learning signal. Removing regularization (ReG\_RL) leads to slower convergence, suggesting it helps accelerate and stabilize the training process. Finally, while BC on the success buffer (ReG\_BC) achieves strong initial performance, RL fine-tuning further improves results, with ReGIL attaining the highest success rates. This demonstrates that retrieval enables effective warm-up and RL refines the policy beyond retrieved trajectories.

\begin{figure*}[tb]
    \centering
    \includegraphics[width=\textwidth]{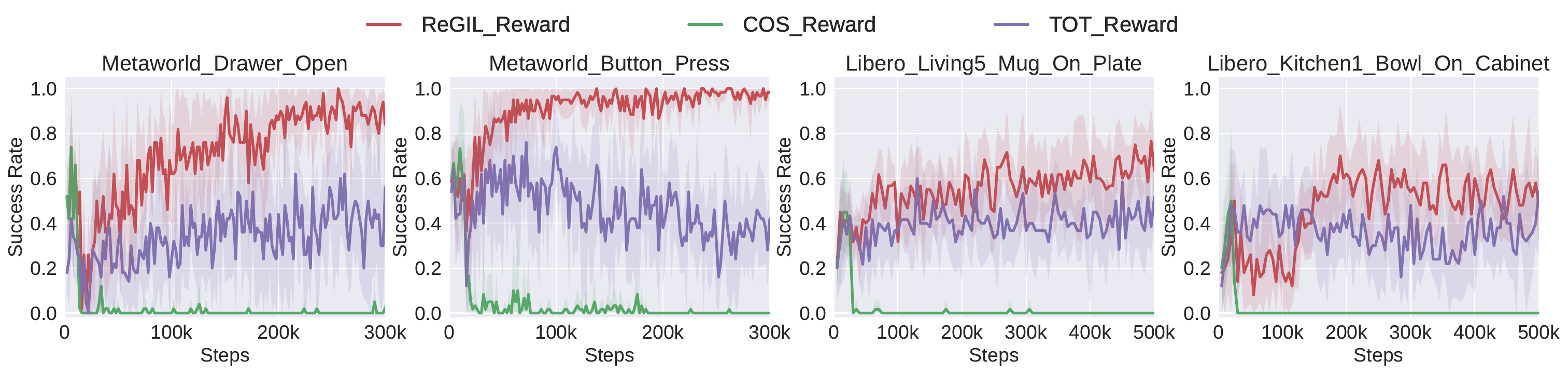} 
    \vspace{-5pt}
\caption{Ablation of Reward Mechanisms.
We compare the proposed reward (ReGIL) against a single-state cosine similarity reward (COS) and the optimal transport-based reward (TOT). The result shows the mean success rate, with the shaded region representing $\pm$1 standard deviation across 5 random seeds and clipped to [0,1]. 
}
 \vspace{-6pt}
\label{fig:ablation_rew}
\end{figure*}

\paragraph{Ablation study on reward} 
ReGIL introduces an observation similarity-based reward derived from retrieved trajectory segments. To evaluate its effectiveness, we compare it with variants that use the same retrieval-based exploration strategy but differ in reward formulation: A state-level reward based on cosine similarity between the current state and a single retrieved state (COS\_Reward), and a trajectory-level reward based on temporally masked OT (TOT\_Reward)~\cite{TOT}. The results in Fig.~\ref{fig:ablation_rew} highlight the importance of segment-level structure. Both TOT\_Reward and the proposed reward significantly outperform COS\_Reward, indicating that state-level similarity is insufficient for guiding optimization. Meanwhile, while TOT\_Reward captures global trajectory alignment, it introduces additional computation complexity and is less responsive to local progress. The proposed retrieval-based reward achieves a better balance by leveraging aligned trajectory segments. Empirically, this advantage is more pronounced on MetaWorld tasks, while on the more complex LIBERO tasks, both methods achieve comparable performance.

\subsection{Results on the Real Robot Tasks}
We compare ReGIL against several baselines, including Behavior Cloning (BC), BC with Retrieval (ReG\_BC), and ReGIL w/o Retrieval (BC\_RL), to evaluate the contribution of the retrieval exploration and retrieve-based reward on the real-world tasks. 

\begin{figure*}[tb]
    \centering
    \includegraphics[width=\textwidth]{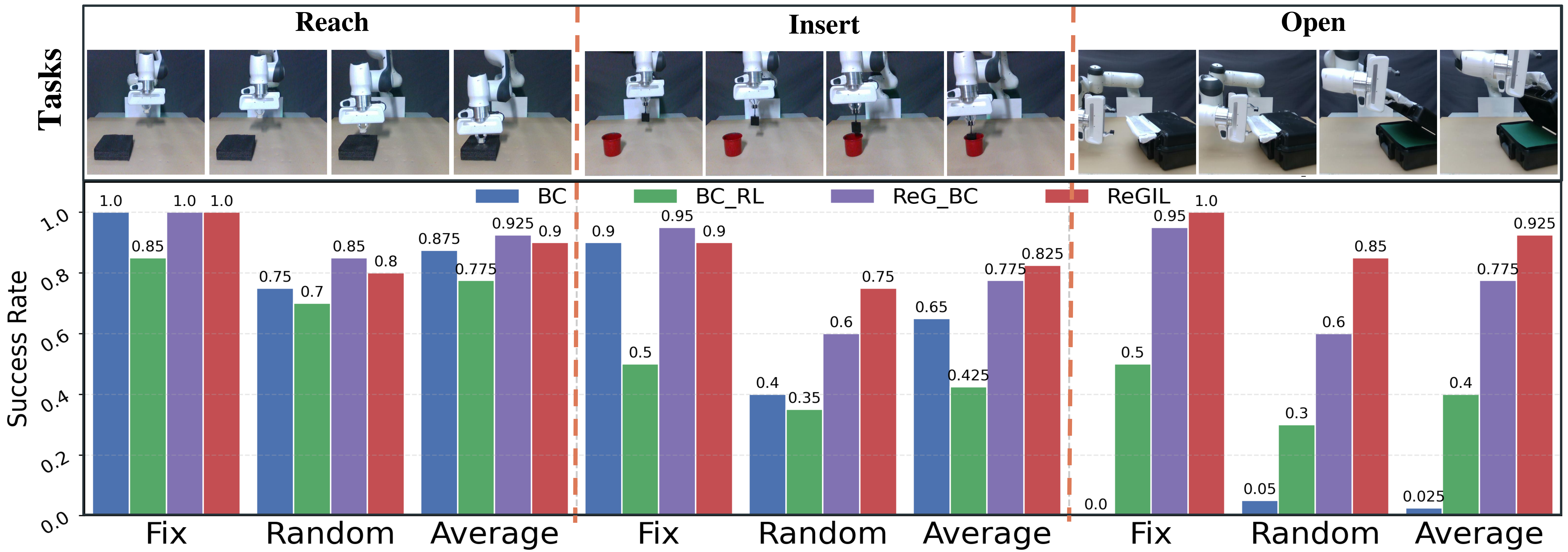} 
    \vspace{-5pt}
\caption{Real robot task visualization and performance comparison. Success rates are reported over 20 trials under fixed-target (Fix), random-target (Random), and the average of both (Average).
}
 \vspace{-5pt}
\label{fig:real_task}
\end{figure*}

\begin{wraptable}[10]{r}{0.63\textwidth}
\small
\centering
\vspace{-18pt}
\caption{Performance comparison across settings. The average success rates under fixed-target ($\text{SR}_{\text{fix}}$) and random-target settings ($\text{SR}_{\text{rand}}$) are reported. p-value is calculated under the random-target setting.}
\label{tab:generalization}
\begin{tabular}{lcccc}
\toprule
Method & $\text{SR}_{\text{rand}}$ $\uparrow$ & $\text{SR}_{\text{fix}}$ $\uparrow$ & $\text{SR}_{\text{rand}} / \text{SR}_{\text{fix}} \uparrow$ & p-value (rand) \\
\midrule
BC      & 40.0\% & 63.3\% & 63.2\% & $< 0.001 $ \\
BC-RL   & 45.0\% & 61.7\% & 72.9\% & $< 0.001 $ \\
ReG\_BC  & \underline{68.3\%} & \textbf{96.7\%} & 70.6\% & 0.144 \\
ReGIL  & \textbf{80.0\%} & \textbf{96.7\%} & \textbf{82.7\%} & -- \\
\bottomrule
\end{tabular}
\end{wraptable}
Real-world results (Fig.~\ref{fig:real_task}) show that both retrieval guidance and online fine-tuning are important. While behavior cloning performs well on simple tasks such as Reach, its performance drops on more complex tasks and under target randomization. Retrieval-guided exploration improves performance by generating additional success trajectories beyond the original demonstration, while online fine-tuning further improves robustness and precision. As shown in Tab.~\ref{tab:generalization}, ReGIL achieves the highest success rate in the random-target setting (80.0\%) while maintaining strong fixed-target performance. Compared with Reg\_BC, ReGIL achieves a higher $\text{SR}_{\text{rand}} / \text{SR}_{\text{fix}}$ ratio, indicating improved robustness beyond simply reproducing trajectories in the success buffer. Overall, these results demonstrate that ReGIL improves generalization and robustness in real-world scenarios. Importantly, the retrieval and reward computation pipeline runs stably above 30Hz, satisfying real-time deployment requirements on real robots.

\section{Discussion}
\label{sec:limitation}
The experimental results suggest that retrieval can play a broader role in one-shot imitation learning. In simulation, ReGIL improves learning efficiency and final performance compared with state-of-the-art baselines. The ablation studies further clarify the contribution of each design choice. In particular, retrieval-guided exploration and the success replay buffer contribute significantly to sample efficiency, especially on long-horizon LIBERO tasks where random exploration rarely reaches meaningful states. The reward ablations further highlight the effectiveness of the proposed reward design, which preserves temporal consistency while remaining efficient enough for online robot learning. Finally, the real-world experiments show that ReGIL enables a more structured and safer exploration strategy, making online adaptation practical in real-world robotic settings.
 
\paragraph{Limitations}  While effective, ReGIL has several important limitations.
First, the method is sensitive to visual ambiguity, particularly in the Insert task, where similar visual patterns can cause retrieval failures and suboptimal guidance, leading to 4 observed failure cases. Performance is also affected by lighting changes, suggesting the need for more robust visual representations. 
A second limitation arises with non-rigid object manipulation, as in the Open task, where handle deformation causes the gripper to slip, resulting in failure. This highlights the limitation of relying solely on visual observations. Incorporating multimodal inputs, such as tactile or contact information, may improve robustness.
In more complex environments, retrieval-guided replay alone could be insufficient to bootstrap successful behavior, reducing the diversity benefit of the success buffer $\mathcal{B}_{\text{suc}}$. Combining retrieval with stronger priors, such as pretrained policies, could mitigate this issue. Future work will also explore adaptive retrieval strategies, further improve the generalization, and leverage action-free demonstrations (e.g., human videos). More details are provided in the Appendix~\ref{app:failure_cases}.


\section{Conclusion}
\label{sec:conclusion}
In this work, we presented ReGIL, a retrieval-guided online imitation learning framework for learning visuomotor manipulation policies from a single demonstration. 
ReGIL treats the demonstration as an external memory during training, allowing retrieval to guide exploration, collect expanded regularization data, and provide dense reward signals. This pipeline improves both data efficiency and robustness under distribution shift. Experiments in both simulation and the real world show that ReGIL outperforms prior baselines in sample efficiency and final task success rate. Notably, ReGIL acquires effective real-world manipulation skills across three tasks using only one demonstration and less than one hour of online interaction. Overall, our results suggest that demonstrations can provide not only static supervision, but also serve as reusable memory throughout the online learning process. We hope this perspective motivates future research on scalable and data-efficient robot learning from limited supervision.

\clearpage
\acknowledgments{The authors acknowledge the use of the MIDAS infrastructure of Aalto School of Electrical Engineering. V. Kyrki acknowledges the research environment provided by ELLIS Institute Finland. This research was supported by the European Union's Horizon Europe research and innovation program under grant agreement No 101189836 (XSCAVE). This research was supported by Business Finland (Aurora, 2479/31/2024).}

\bibliography{ref}  

\renewcommand{\thesubsection}{\Alph{subsection}}

\section*{Appendices}
\label{app:appendix}

\subsection{ReGIL Algorithm Detail}
\label{app:alg}

\subsubsection{Overall Algorithm Detail}
\label{app:alg_overall}
ReGIL is an online imitation learning method that leverages retrieval-guided exploration and optimization from demonstrations. At each timestep, the current observation history is temporally aligned with the demonstration trajectory to retrieve the most relevant expert segment. During an initial exploration phase, the agent follows retrieved expert actions to efficiently discover successful behaviors. Successful trajectories collected during this phase are stored as an expanded memory buffer, which provides additional regularization data beyond the original demonstrations. After warm-up, the learned policy takes control while retrieval continues to provide dense supervision through trajectory alignment. To improve efficiency, retrieval is performed only once after observing the next state, and the resulting alignment is reused for both reward computation and subsequent action selection. At the beginning, the history window $ H $ gradually grows. We initialize $\mathcal{B}_{\text{suc}}$ with the original demonstration $\tau^e$ so that the BC regularizer in Eq.~\ref{eq:policy_opt} is well-defined throughout training. Successful trajectories collected during the retrieval-guided warm-up are then appended to $\mathcal{B}_{\text{suc}}$, progressively enriching the regularization data with online experience. In the worst case where no warm-up trajectory succeeds, the method reduces to TD3 with the retrieval-based reward and BC regularization on the single demonstration. 

Moreover, the reward proxy admits a basic consistency with the exploration phrase. The expert trajectory is a fixed point of the reward signal: an agent that perfectly replays the demonstration receives the maximum reward at every step. The overall training procedure is summarized in Algorithm~\ref{alg:regil}, and the hyperparameters are listed in Tab.~\ref {tab:hyperparameters}.
 
\begin{algorithm}[h]
\caption{ReGIL: Retrieval-Guided Imitation Learning}
\label{alg:regil}
\begin{algorithmic}[1]

\REQUIRE Demonstration $\tau^e$, Warm-up steps $T_w$, total training steps $T$ 
\STATE Initialize policy $\pi_\theta$, Q-functions $Q_{\phi_1}, Q_{\phi_2}$,empty buffers $\mathcal{B}$ \STATE Initialize $\mathcal{B}_{\text{suc}}$ with $ \tau^e$
\STATE Environment reset $o_0=\textit{env.reset()}$
\STATE Retrieve demonstration segment$\tau^e_{j_0^{\text{start}}:j_0^*}$ by local temporal alignment

\FOR{$t = 1$ to $T$}
    \IF{$t < T_w$}
        \STATE $a_t \leftarrow a^e_{j_t^*-1}$ \hfill $\triangleright$ Retrieval-guided exploration
        
    \ELSE
        \STATE Sample $a_t \sim \pi_\theta(\cdot \mid o_t)$     \hfill $\triangleright$ Policy action
    \ENDIF

    \STATE $o_{t+1}, \text{done}, \text{success} = \textit{env.step}(a_t) $ \hfill $\triangleright$ Execute action

    \STATE Align $\tau_{t-H+1:t+1}$ with $\tau^e_{0:q_{t+1}^*}$ to obtain $ j_{t+1}^*$ \hfill $\triangleright$ Reuse $j_{t+1}^*$ for next-step action $a_t$

    \STATE Compute reward $r_{t+1}$ according to Eq.\ref{eq:reward}

    \STATE Store $(o_t, a_t, r_{t+1}, o_{t+1},\text{done})$ in $\mathcal{B}$
    \IF{done}
        \STATE Environment reset $o_0=\textit{env.reset()}$
        \STATE Retrieve demonstration segment$\tau^e_{j_0^{\text{start}}:j_0^*}$ by local temporal alignment
        \IF{success \textbf{and} $t < T_w$}
            \STATE Store trajectory in $\mathcal{B}_{\text{suc}}$
        \ENDIF
    \ENDIF
    \STATE Sample batch from $\mathcal{B}$ and $\mathcal{B}_{\text{suc}}$
    \STATE Update $Q_{\phi_1}, Q_{\phi_2}$ and  $\pi_\theta$ according to Eq.\ref{eq:policy_opt} \hfill $\triangleright$ Update critic and actor

\ENDFOR

\STATE \textbf{return} policy $\pi_\theta$

\end{algorithmic}
\end{algorithm}

\subsubsection{Reward Design Comparison}
\label{app:reward_comparison}

\begin{figure*}
    \centering
    \includegraphics[width=\textwidth]{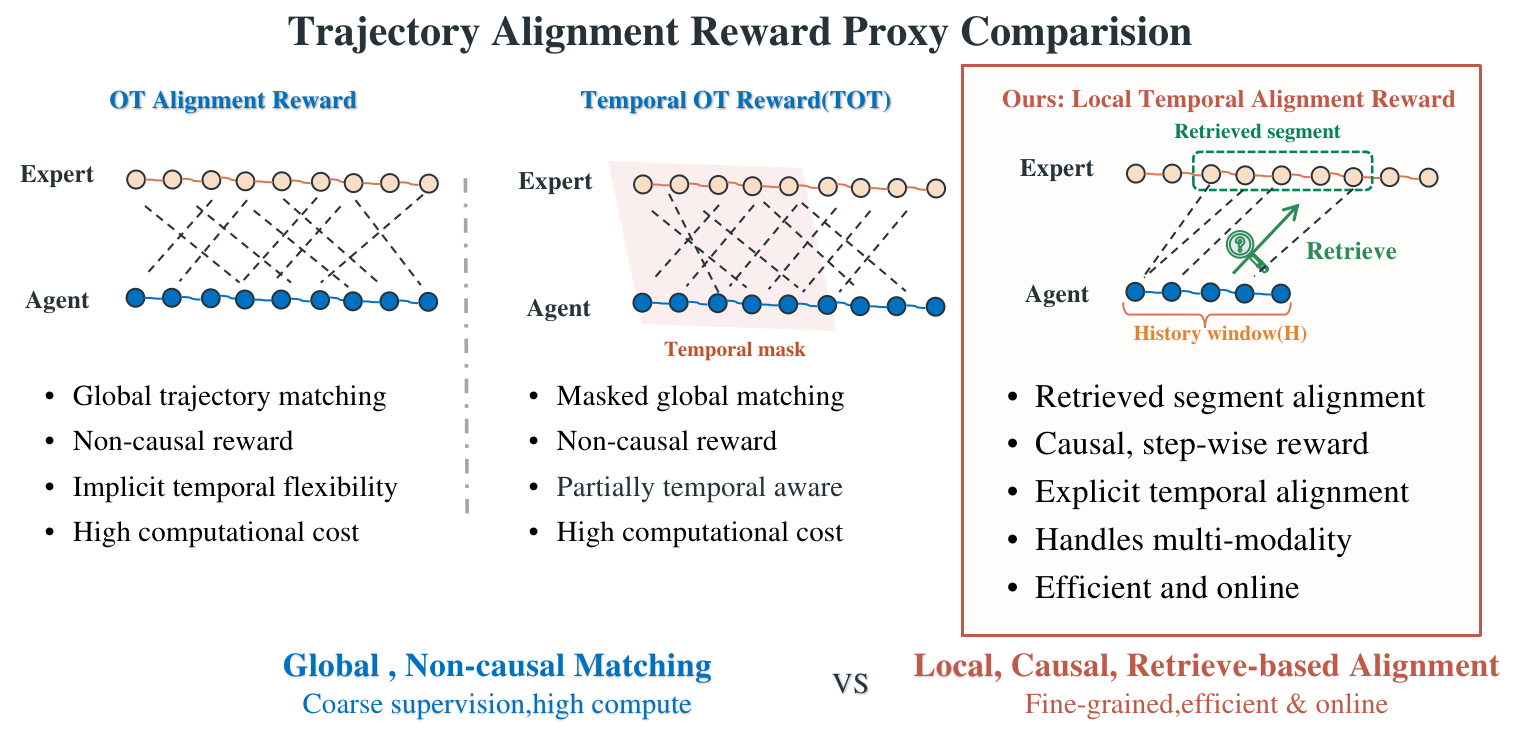} 
\caption{\textbf{Comparison of trajectory alignment reward formulations.} 
 We compare three classes of reward design: (left) global optimal transport (OT) alignment, (middle) temporally masked OT (TOT), and (right) our proposed local retrieval-based temporal alignment.
}
\label{fig:reward}
\end{figure*}

We compare the proposed reward with the other two rewards based on the observation alignment principle, including optimal transport (OT) alignment and temporally masked OT. Fig.\ref{fig:reward} highlights the fundamental differences between global OT-based rewards and our proposed local retrieval-based formulation. We emphasize that this is a trade-off rather than a strict improvement. In addition, although we do not evaluate multiple demonstration settings in the present work, we note that the local-alignment formulation in principle can extend to this setting without any modification or easily use Faiss to enable retrieval in seconds.

The $1/(H{+}1)$ factor in Eq.~(\ref{eq:reward}) normalizes by the agent window length, not by the alignment path length (which depends on the warping). This keeps $r_{t+1}$ on a comparable scale across timesteps once $H$ is fixed and encourages the shortest path selection.

\subsubsection{TD3 Algorithm Detail}
We employ an adapted TD3 algorithm for policy optimization. The actor predicts the mean action $\mu_{\theta}(o)$, while fixed Gaussian noise is used during training for exploration and likelihood-based imitation regularization $\pi_\theta(a_t \mid o_t)=\mathcal{N}\big(\mu_{\theta}(o_t), \delta^2 I\big),$
where $\delta$ is a predefined constant. During evaluation, we execute the mean action $\mu_\theta(o_t)$, yielding deterministic behavior.

We maintain two Q networks $Q_{\phi_1}$ and $Q_{\phi_2}$, and define $Q_{\min}(o,a)=\min\{Q_{\phi_1}(o,a),Q_{\phi_2}(o,a)\}$
and online replay buffer $\mathcal{B}$. The Q function objective is:
\begin{equation}
\mathcal{L}_Q(\phi_i)=\mathbb{E}_{(o_t,a_t,r_{t+1},o_{t+1})\sim \mathcal{B}}\big[(Q_{\phi_i}(o_t,a_t)-(r_{t+1}+\gamma Q_{\min}\big(o_{t+1},\pi_{\bar{\theta}}(o_{t+1})))
)^2\big].
\label{eq:q_loss}
\end{equation}
The actor is optimized using Eq.\ref{eq:policy_opt} with a linear decay schedule $ \lambda(t)=\lambda_0 \max(1-t/T_{bc},0) $ which decays over training steps $t$. We note that once $\lambda(t)=0$, the actor objective reduces to pure $Q$-maximization under the retrieval-based reward, so the policy depends on the fidelity of that reward signal.

\begin{table}
\centering
\caption{Hyperparameters for ReGIL}
\label{tab:hyperparameters}
\small
\begin{tabular}{lll}
\toprule
\textbf{Hyperparameter} & \textbf{Value} & \textbf{Description} \\
\midrule
\texttt{lr}  & 0.00005 & Learning rate\\
\texttt{gamma} & 0.99 & Reward discount horizon factor \\
\texttt{policy\_freq} & 2 & Delayed actor update frequency\\
\texttt{encoder\_type} & \texttt{resnet} & Visual backbone \\
\texttt{obs\_shape}& $84 \times 84$ for real world & Input dimensions \\
\texttt{action\_shape}&$4$ for real world & Dimension of action space\\
\texttt{batch\_size} & $128$ for libero , $64$ for metaworld & Mini-batch size for optimization updates \\
\texttt{warm-up} & $20k$ for libero , $5k$ for metaworld & The retrieve-guided exploration step number \\
\texttt{suc\_buff\_size} & $10k$ for libero , $5k$ for metaworld & The capability of success buffer \\
\bottomrule
\end{tabular}
\vspace{-5pt}
\end{table}

\subsection{Experimental Details}
\subsubsection{Simulation Experimental Details}
\label{app:simulation}
\paragraph{Data processing}
For data processing, we limited all tasks in Metaworld to the first 300k training steps and tasks in Libero to the first 500k steps with a 10k rolling window for the online learning methods. Both behavior cloning and BAKU are trained for 100k steps.

\paragraph{Baseline selection and additional Result}
ReGIL is a retrieval-based imitation learning method. Therefore, we compare it against several representative imitation and trajectory-matching approaches under the same single-demonstration setting. The selected baselines cover both standard offline imitation learning and state-of-the-art reward-based online imitation frameworks, as listed below:

\begin{itemize}[nosep]
    \item \textbf{Behavior Cloning (BC)}: A standard imitation learning method that directly learns a policy by supervised learning on expert state–action pairs.
    \item \textbf{BAKU}: A transformer-based behavior cloning method that models sequential dependencies using an action-chunking strategy\citec{baku}.
    \item \textbf{Regularized Trajectory-matching IRL (ROT)}: A method that combines trajectory-matching rewards with behavior cloning pre-training and regularization to accelerate imitation learning in the low-demonstration regime\citec{ROT}. 
    \item \textbf{Temporal Trajectory-matching IRL (TOT)}: A state-of-the-art trajectory-matching method that incorporates temporal mask and context information to learn a more accurate optimal-transport-based proxy reward\citec{TOT}.
\end{itemize}

\begin{figure*}[tb]
    \centering
    \includegraphics[width=0.9\textwidth]{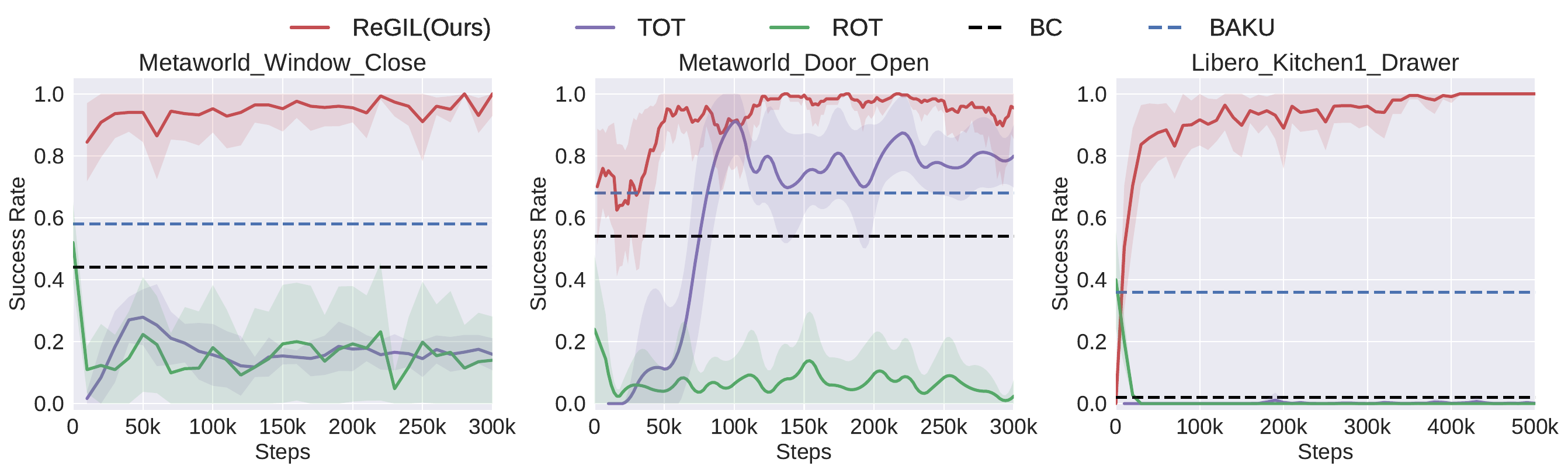} 
    \vspace{-5pt}
\caption{
Additional Baseline Comparison in Simulation. All methods are trained with a single demonstration on Meta-World and LIBERO. Curves show mean success rate, with shaded regions indicating $\pm$1 standard deviation across five seeds, clipped to [0, 1].
}
\label{fig:baseline_additional}
\end{figure*}

The additional baseline comparison is shown in Fig.~\ref{fig:baseline_additional}. ReGIL consistently outperforms all baselines across both Meta-World and LIBERO benchmarks, demonstrating improved sample efficiency and stronger final performance under the single-demonstration setting. 

\paragraph{Ablation study details}
To better understand the contribution of each component in ReGIL, we conduct ablation studies based on the following ablated variants:
\begin{itemize}[nosep]
    \item \textbf{ReGIL w/o Retrieval (BC\_RL)}: RL with BC regularization using only the single expert demonstration, without retrieval-based exploration. This variant evaluates the importance of the retrieval mechanism.
    \item \textbf{ReGIL w/o Regularization (ReG\_RL)}: RL with retrieval-based data collection but without BC regularization on success buffer $ \mathcal{B}_{\text{suc}}$. This variant tests whether BC regularization is necessary to stabilize policy learning.
    \item \textbf{RL with Random Exploration (RL)}: A standard RL baseline with random exploration, in other words, ReGIL w/o both Retrieval and Regularization. This evaluates the benefit of guided exploration and bc regularization.
    \item \textbf{ReGIL w/o Optimization (ReG\_BC)}: Behavior cloning trained on the success buffer without RL fine-tuning. This variant evaluates whether reinforcement learning further improves the performance.
\end{itemize}

\subsubsection{Real-World Experimental details}
\label{app:real_world}
\paragraph{Real-World Environment Setup}
\begin{wrapfigure}[13]{r}{0.45\textwidth}
    \centering
    \vspace{-10pt}
    \includegraphics[width=0.42\textwidth]{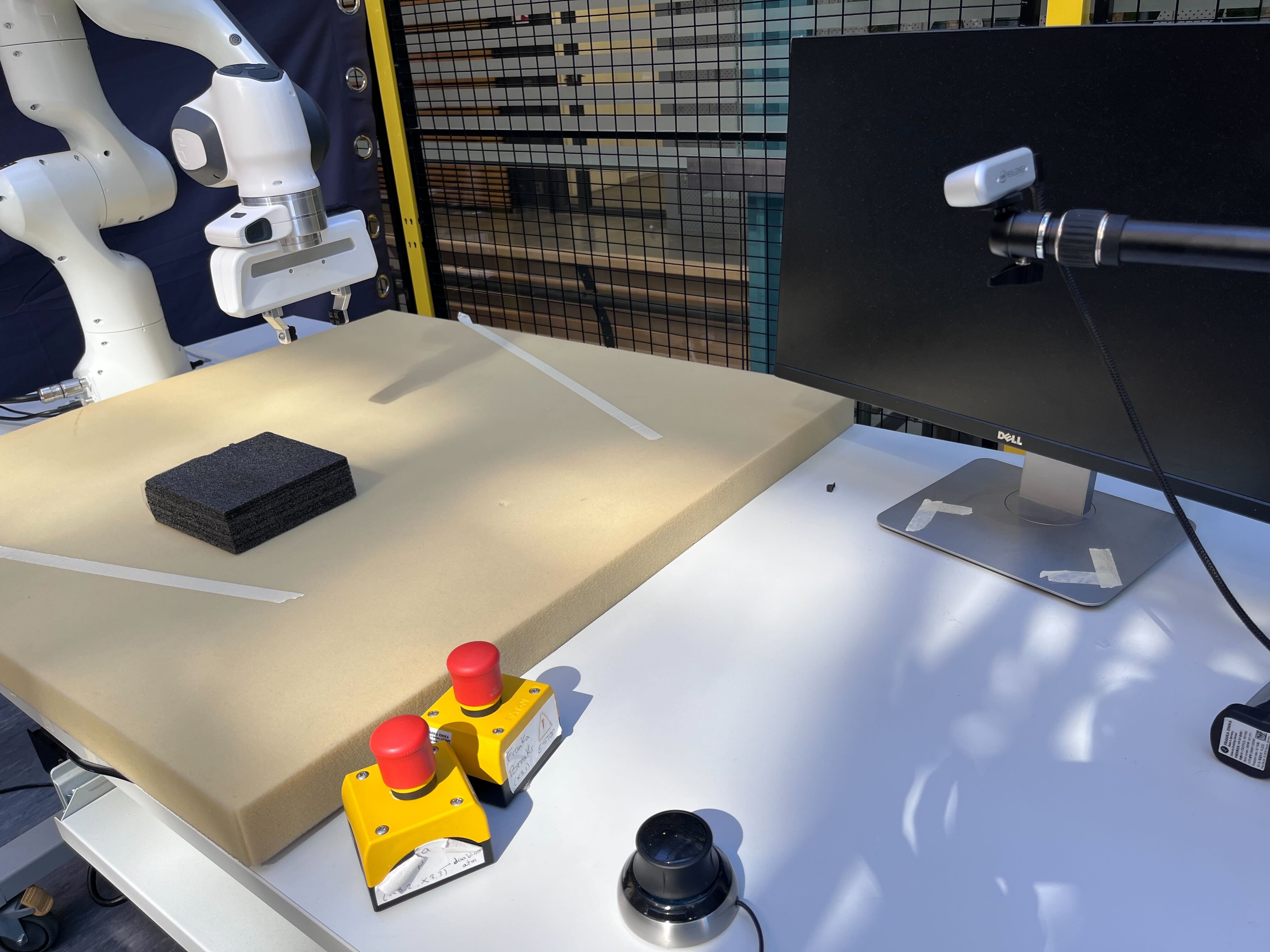}  
\caption{\textbf{Real-world experimental setup.} 
}
\label{fig:task_setup}
\end{wrapfigure}
We evaluate ReGIL on a Franka Emika Panda robot equipped with an Intel RealSense D435 RGB camera. The robot is controlled via Cartesian delta actions $(dx, dy, dz)$ with a binary gripper command at a nominal frequency of approximately 1 kHz. Visual observations consist of $84 \times 84$ RGB images captured at 30 Hz from a fixed external viewpoint. A single demonstration is collected via a 3D connexion SpaceMouse through manually teleoperation. Although low-level control runs at 1 kHz, the closed-loop system is constrained by the 30 Hz RGB camera stream. Retrieval and reward computation are executed above 30 Hz( shown in Table~\ref{tab:computational_efficiency}), ensuring real-time performance for robotic deployment.

Fig.\ref{fig:task_setup} illustrates our real-world experimental setup. The robot receives only RGB observations from a fixed external camera, without access to depth, proprioceptive state estimation, or privileged information. Actions are specified in Cartesian space through delta end-effector commands, making the control problem continuous.

\paragraph{Real World Task Descriptions}
We design three distinct tasks to evaluate the performance of approaches and use a single RGB as input, therefore we limited the changes along with the camera direction.

\textbf{Reach}: The robot must move its end effector to contact the black target region before the maximum episode length is reached. Two online agents are trained for 8k training steps, with 1000 guided exploration steps, around 40 minutes wall clock.  For this task, the max step is 125. For the random target evaluation, the position of the target will move in [-14cm,8cm] in left and right direction, given the fix-target position as the original point.

\textbf{Insert}: The robot must insert the black rigid plug attached to the end of a cable into the red cup. A trajectory is considered successful when the bottom of the plug is inside the cup. For this task, the max step is 150. Two online agents are trained for 10k training steps, with 1300 guided exploration steps, around 1 hours wall clock. For the random target evaluation, the position of the cup will move in [-4.5cm,4.5cm] in left and right direction, given the fix-target position as the original point.

\textbf{Open}: The robot must open the black box while keeping its base approximately stationary. The robot's end-effector must establish and maintain specific contact with the box lid to overcome friction and hinge constraints.  For this task, the max step is 175. Two online agents are trained for 8k training steps, with 1300 guided exploration steps, around 40 mins wall clock each. For the random target evaluation, the position of the cup will move in [-4.0cm,7.0cm], given the fix-target position as the original point.

\subsection{Sensitivity Study on Retrieval Parameters}
\label{app:exp_params_study}

\begin{table}[h]
    \centering
    \small
    \setlength{\tabcolsep}{6.5pt} 
    \begin{tabular}{l|ccc|cccc|ccc}
        \toprule
        & \multicolumn{3}{c|}{\textbf{History Length (H) [k=5]}} & \multicolumn{4}{c|}{\textbf{Candidate size (k) [H=4]}} & \multicolumn{3}{c}{\textbf{Encoder [k=5, H=4]}} \\
        \textbf{Benchmark} & \textbf{H=9} & \textbf{H=4} & \textbf{H=2} & \textbf{k=10} & \textbf{k=5} & \textbf{k=3} & \textbf{k=1} & \textbf{CLIP} & \textbf{DINO} & \textbf{R3M} \\
        \midrule
        MetaWorld & 0.48 & 0.53 & 0.46 & 0.53 & 0.53 & 0.50 & 0.48  & 0.475 & 0.53 & 0.45 \\
        LIBERO & 0.50 & 0.47 & 0.42 & 0.47 & 0.47 & 0.50 & 0.37  & 0.017 & 0.47 & 0.17 \\
        
        \midrule
        \textbf{Mean} &0.49 & \textbf{0.50} & 0.44 & \textbf{0.50} & \textbf{0.50} & \textbf{0.50} & 0.425 & 0.246 & \textbf{0.50} & 0.31 \\
        \bottomrule
    \end{tabular}
    \vspace{5pt}
    \caption{
    \textbf{Sensitivity analysis of retrieval design choices.} 
    We evaluate the impact of history length $H$, candidate size $k$, and visual encoder selection on MetaWorld and LIBERO. 
    }
    \vspace{-5pt}
    \label{tab:param_retrieve}
\end{table}

We study the impact of three key design choices for the retrieval process, including history length $H$, candidate size $k$, and visual encoder selection. 
Notably, when a history window of length $H$ is used, retrieval is conditioned on $H+1$ consecutive states, including the current timestep. Experiments are conducted on MetaWorld and the LIBERO task suite, reporting the average success rate on 50 trials each. The result is shown in Tab.\ref{tab:param_retrieve}. In general, the visual encoder choice has a substantial impact. DINOv3\citec{dinov3} consistently outperforms CLIP\cite{clip} and R3M\citec{r3m} across both benchmarks. For the history length, using a moderate history ($H=4$) yields the best overall performance, while shorter histories ($H=2$) lack sufficient temporal context. Finally, we observe that performance is stable for $k \in {3,5,10}$, indicating robustness to this hyperparameter. However, using only a single nearest neighbor ($k=1$) significantly degrades performance, suggesting that multiple candidates are necessary to avoid incorrect retrievals. Overall, these results suggest that ReGIL is robust to moderate variations in retrieval parameters, while benefiting from temporally aware alignment and strong visual representations. For the experiment in this work, we use $H=4, k=5$ and the frozen DINOv3 encoder for all experiments.

\begin{figure*}[tb]
    \centering
    \includegraphics[width=\textwidth]{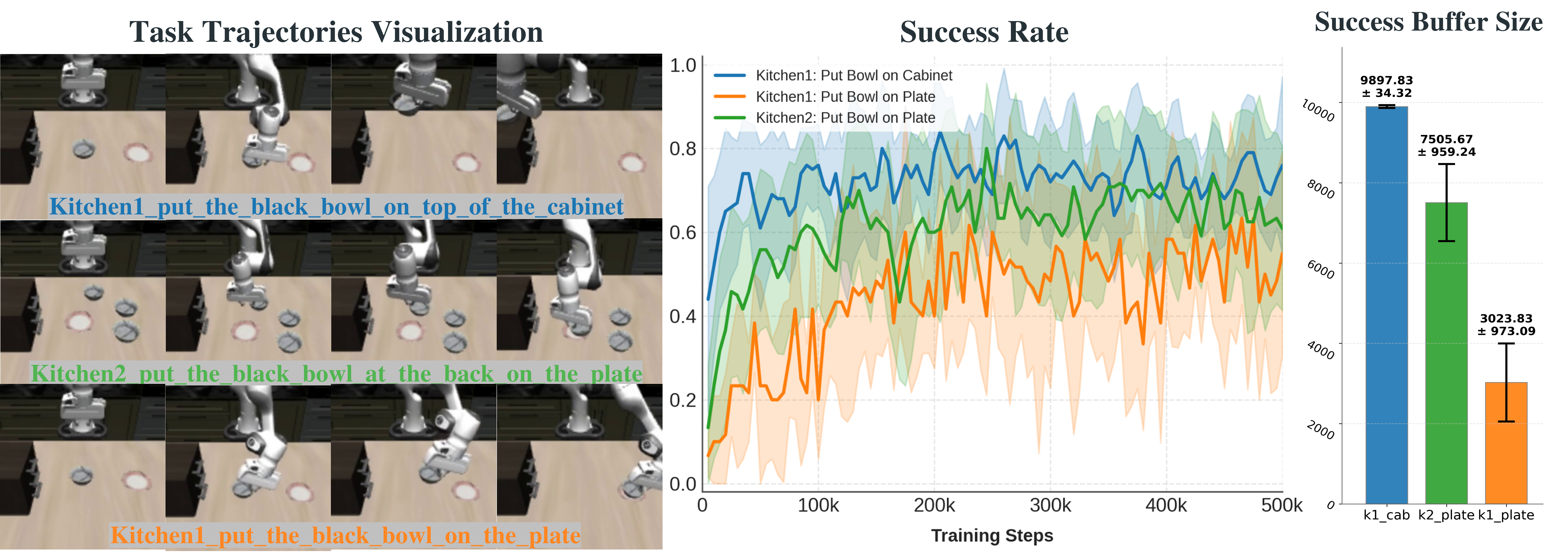} 
    \vspace{-5pt}
\caption{
Task trajectory visualization and training performance comparison across three similar kitchen manipulation tasks. Left: Representative task execution trajectories from three long-horizon robotic manipulation tasks. Middle: Success rates across 5 runs, and the shaded regions indicate the standard deviation. Right: Average success buffer size collected during training.
}

\label{fig:sf_com}
\vspace{-5pt}
\end{figure*}

\begin{table}[h]
\centering
\caption{Success buffer size across tasks.
We report the mean and standard deviation of successful trajectories collected in the success buffer $\mathcal{B}_{\text{suc}}$ for each task across simulation and real-world environments. The maximum buffer capacity is indicated for each environment suite.
}
\label{tab:sf_size}
\begin{tabular}{cccc}
\toprule
\textbf{Environment} & \textbf{Suite} & \textbf{Task Name} & \textbf{Success Buffer Size} \\ 
 & (Maximum)& & (Mean $\pm$ Std) \\
\midrule
\multirow{12}{*}{\textbf{Simulation}} 
 & \multirow{7}{*}{MetaWorld } 
   & button\_press     & 1716.00 $\pm$ 266.17 \\
 & & door\_open        & 2841.50 $\pm$ 674.94 \\
 & & drawer\_close     & 3222.80 $\pm$ 244.98 \\
 & & drawer\_open      & 2301.00 $\pm$ 243.07 \\
& (5000)&\cellcolor[HTML]{FFE4E1}hammer & \cellcolor[HTML]{FFE4E1}455.83 $\pm$ 146.58 \\
 & & window\_close     & 3031.20 $\pm$ 544.19 \\
 & & window\_open      & 1807.60 $\pm$ 500.73 \\
\cmidrule{2-4}
 & \multirow{5}{*}{LIBERO} 
   & kitchen1\_bowl\_on\_cabinet    & 9897.83 $\pm$ 34.32  \\
 & & kitchen1\_drawer  & 9915.80 $\pm$ 22.56  \\
 & & kitchen1\_bowl\_on\_plate    & 3023.83 $\pm$ 973.09 \\
 & & kitchen2\_bowl\_on\_plate    & 7505.67 $\pm$ 959.24 \\
 & (10000)& living5\_mug\_on\_plate          & 6228.40 $\pm$ 765.61 \\
 & & study4           & 9961.17 $\pm$ 25.16  \\
\midrule
\multirow{3}{*}{\textbf{Real World}} 
 & 1000& reach            & 814.00               \\
 & 1300& insert           & 961.00              \\
 & 1300& open             & 947.00              \\
 
\bottomrule
\end{tabular}
\end{table}

\subsection{Success buffer Analysis}
\label{app:sf_comparision}

\subsubsection{Diversity Analysis}
\begin{figure*}[tb]
    \centering
    \includegraphics[width=0.65\textwidth]{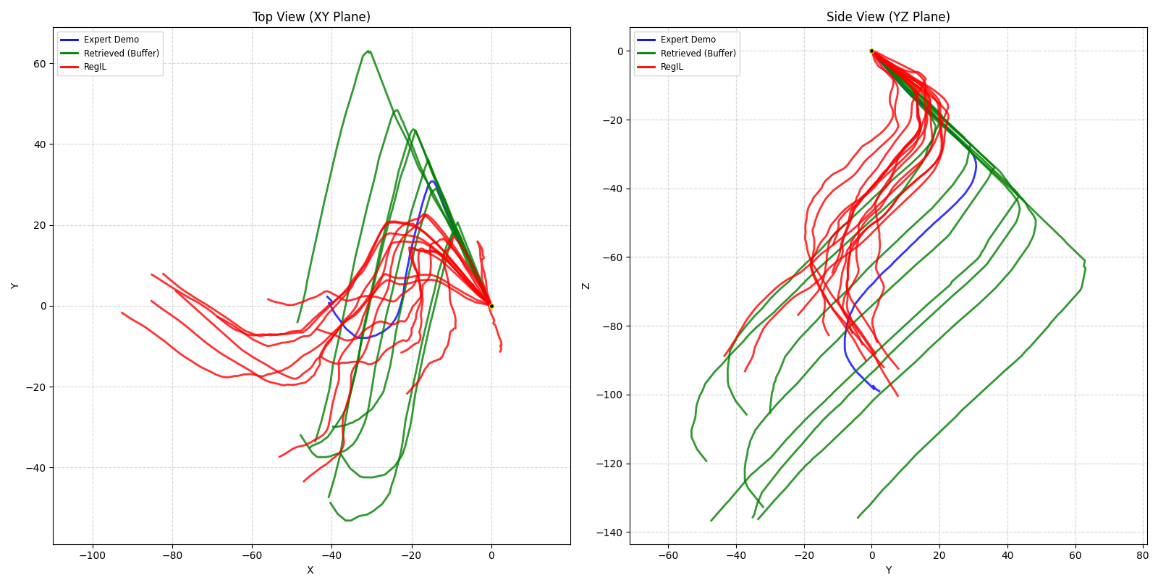} 
    \vspace{-5pt}
\caption{End-effector trajectory comparison on the Open task.
We visualize the expert demonstration (blue), trajectories stored in the success buffer (green), and trajectories generated by the final ReGIL policy (red). 
The expert dataset contains a single trajectory corresponding to a fixed execution strategy. In contrast, retrieval-guided exploration produces a diverse set of successful trajectories that significantly expands coverage of the state space. The final learned policy further refines these behaviors, generating smoother and more consistent trajectories that generalize across varying initial conditions.
}
\vspace{-5pt}
\label{fig:traj_compare}
\end{figure*}

To illustrate the diversity of successful trajectories stored in the success buffer, we visualize the end-effector trajectories of the expert demonstration, the success buffer, and the final ReGIL policy in Fig.~\ref{fig:traj_compare}. During evaluation, the initial end-effector pose is randomized for each episode. For visualization purposes, all trajectories are translated to a common starting position, allowing a direct comparison of their geometric structures.

The expert data contains only a single trajectory corresponding to one specific execution strategy. In contrast, the retrieval-guided exploration phase collects multiple successful trajectories that exhibit substantial variation in their motion patterns. This indicates that retrieval enables structured exploration and allows the agent to discover successful behaviors beyond those covered by the original demonstration distribution. 

Furthermore, the trajectories generated by the final ReGIL policy differ from both the original demonstration and the retrieved trajectories.  These results suggest that reinforcement learning, combined with retrieval-extended regularization, enables the policy to exploit the diverse successful experiences stored in the buffer and learn a more generalizable solution.

\subsubsection{Quality Analysis}

We analyze the number of successful samples collected during the retrieval-guided exploration phase, as summarized in Tab~\ref{tab:sf_size}. Across most tasks, ReGIL accumulates a substantial number of successful samples. However, the hammer task stands out with a markedly reduced number of successful trajectories (455.83 ± 146.58), indicating that this task is significantly more difficult for retrieval-guided exploration. This aligns with the observed performance gap in the main experiments, suggesting a correlation between success buffer size and downstream policy performance.

To further investigate this effect, we compare three semantically similar tasks from the LIBERO benchmark, shown in Fig.~\ref{fig:sf_com}: kitchen1\_put\_bowl\_on\_cabinet, kitchen1\_put\_bowl\_on\_plate, and kitchen2\_put\_bowl\_on\_plate. While all tasks require similar manipulation skills, the successful trajectories collected during exploration are different. Consistent with this observation, the corresponding environment with fewer samples exhibits slower performance improvement throughout training. This trend suggests that the quantity of successful trajectories in $\mathcal{B}_{\text{suc}}$ influences the effectiveness of retrieval-based regularization. A larger and more diverse success buffer provides richer supervision signals, facilitating faster policy improvement, whereas limited successful experience can reduce the benefit of retrieval guidance and slow convergence.

These results highlight the importance of retrieval quality for ReGIL. Accurate retrieval facilitates the discovery of successful trajectories during exploration, resulting in a larger and more informative success buffer. The retrieved experiences subsequently provide stronger guidance for reward computation and policy regularization during policy optimization.

\subsection{Computational Efficiency Analysis}
\label{app:compute_efficiency}

\begin{table}[htbp]
\centering
\caption{Latency Breakdown and Computational Efficiency}
\label{tab:computational_efficiency}
\begin{tabular}{lcccc}
\toprule
\textbf{Task} & \textbf{\begin{tabular}[c]{@{}c@{}}Total Pipeline\\ Cost (ms)\end{tabular}} & \textbf{\begin{tabular}[c]{@{}c@{}}Pure DINO\\ Encoding (ms)\end{tabular}} & \textbf{\begin{tabular}[c]{@{}c@{}}Pure Search +\\ S-DTW (ms)\end{tabular}} & \textbf{\begin{tabular}[c]{@{}c@{}}Max Control\\ Freq. (Hz)\end{tabular}} \\ \midrule
MetaWorld& $31.61 \pm 2.17$ & $30.30 \pm 1.84$ \scriptsize{(95.9\%)} & $1.29 \pm 0.43$ \scriptsize{(4.1\%)} & $31.64$ \\
LIBERO & $31.07 \pm 1.04$ & $29.61 \pm 1.02$ \scriptsize{(95.3\%)} & $1.44 \pm 0.05$ \scriptsize{(4.6\%)} & $32.18$ \\ \midrule
Real Insert          & $32.85 \pm 4.64$ & $30.56 \pm 3.22$ \scriptsize{(93.0\%)} & $2.26 \pm 2.38$ \scriptsize{(6.9\%)} & $30.45$ \\
Real Reach           & $31.94 \pm 3.92$ & $30.09 \pm 3.82$ \scriptsize{(94.2\%)} & $1.83 \pm 0.21$ \scriptsize{(5.7\%)} & $31.31$ \\
Real Open                  & $31.44 \pm 1.32$ & $29.32 \pm 1.29$ \scriptsize{(93.3\%)} & $2.09 \pm 0.21$ \scriptsize{(6.7\%)} & $31.81$ \\ \bottomrule
\end{tabular}
\end{table}

To isolate the computational bottlenecks of our framework, we conduct a latency profiling. The evaluation benchmarks our pipeline across two simulation suites (window close in MetaWorld and bowl on the cabinet for LIBERO) and three real robot tasks (Insert, Reach, and Open) with 150 steps. All trials are executed under identical hyperparameters ($H=4, k=5$).

As comprehensively compiled in Table~\ref{tab:computational_efficiency}, the total end-to-end processing pipeline exhibits tight numerical convergence across all evaluation tokens, consistently falling within the narrow range of $31.07\sim32.85$\,ms per step. This translates to an empirical operational speed of over $30$\,Hz ($\sim31.5$\,Hz on average), which satisfies the real-time closed-loop requirement of high-frequency visual imitation learning. 

Crucially, our structured time breakdown explicitly delineates the source of computational overhead. Visual feature extraction through the pre-trained DINOv3 backbone dominates the processing cycle, consuming $93.0\%\sim95.9\%$ of the entire pipeline latency. In contrast, our key algorithmic contribution proves to be lightweight. Across both simulated benchmarks and empirical real-world setups, the pure search and alignment phase requires an average of merely $1.29\sim2.26$\,ms, accounting for a negligible fraction ($4.1\%\sim6.9\%$) of the computational budget.

\subsection{Failure Cases Analysis}
\label{app:failure_cases}

\begin{figure*}[tb]
    \centering
    \includegraphics[width=0.95\textwidth]{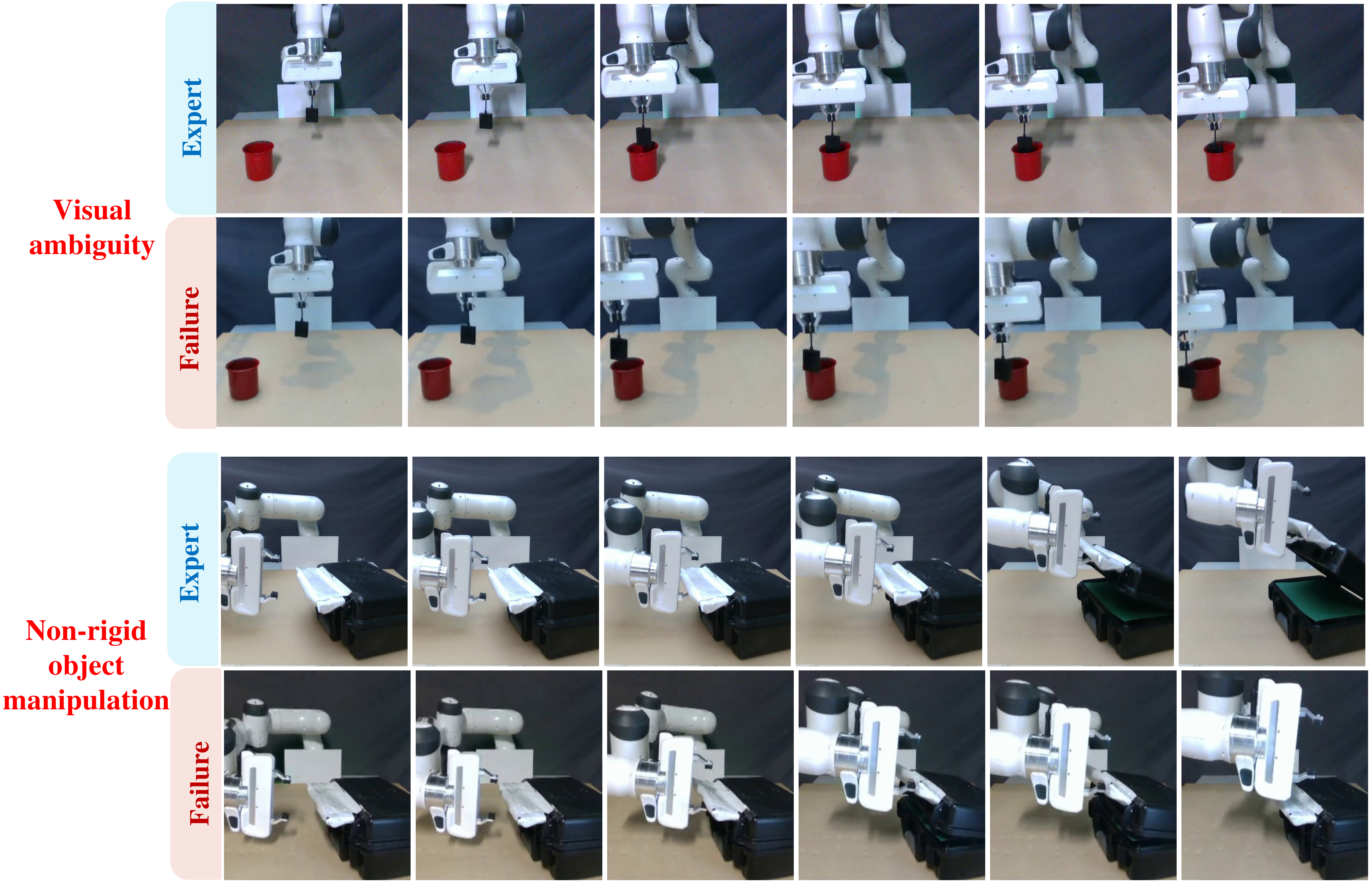} 
\caption{\textbf{Expert Execution versus a typical Failure Case.} 
We visualize the expert trajectory and failed executions for both the Insert task (top) and the Open task (bottom).
}
\vspace{-8pt}
\label{fig:failure}
\end{figure*}

While ReGIL achieves strong performance across diverse manipulation tasks, our real-world experiments reveal several representative failure modes: (i) visual ambiguity in the Insert task, where visually similar configurations confuse retrieval and control, resulting in 4/4 failures; (ii) limitations in non-rigid object manipulation, where vision-only input makes it difficult to maintain stable contact and apply appropriate forces to deformable handles which is a common constraint of vision-only policies, leading to failures in all Open-task trials(shown in Fig.~\ref{fig:failure}).; and (iii) policy errors in the Reach task, where failures stem primarily from control, indicating headroom to further improve generalization.


\end{document}